\documentclass[sigconf, authorversion]{acmart}
\makeatletter
\renewcommand\@formatdoi[1]{\ignorespaces}
\makeatother
\renewcommand\footnotetextcopyrightpermission[1]{} 
\usepackage{balance} 

\usepackage[linesnumbered,ruled,vlined]{algorithm2e}
\usepackage{tablefootnote}

\usepackage{booktabs} 
\usepackage{balance,url}
\usepackage{color,soul}
\usepackage{graphicx}
\usepackage{multirow}
\usepackage{array}
\usepackage{dsfont}
\usepackage{makecell}
\usepackage{caption}
\usepackage{subcaption}

\usepackage{amsmath,amsfonts}

\usepackage{graphicx}
\usepackage{balance}

\begin{document}

\title{Energy-Efficient Parking Analytics System using Deep Reinforcement Learning}

\author{Yoones Rezaei}
\email{yor10@pitt.edu}
\affiliation{
  \institution{University of Pittsburgh, Pittsburgh}
}
\author{Stephen Lee}
\email{stephen.lee@pitt.edu}
\affiliation{
  \institution{University of Pittsburgh, Pittsburgh}
}
\author{Daniel Mosse}
\email{mosse@pitt.edu}
\affiliation{
  \institution{University of Pittsburgh, Pittsburgh}
}
\begin{abstract}
Advances in deep vision techniques and ubiquity of smart cameras will drive the next generation of video analytics. 
However, video analytics applications consume vast amounts of energy as both deep learning techniques and cameras are power-hungry. In this paper, we focus on a parking video analytics platform and propose RL-CamSleep, a deep reinforcement learning-based technique, to actuate the cameras to reduce the energy footprint while retaining the system's utility. Our key insight is that many video-analytics applications do not always need to be operational, and we can design policies to activate video analytics only when necessary. Moreover, our work is complementary to existing work that focuses on improving hardware and software efficiency. 
We evaluate our approach on a city-scale parking dataset having 76 streets spread across the city. Our analysis demonstrates how streets have various parking patterns, highlighting the importance of an adaptive policy. Our 
approach can learn such an adaptive policy that can reduce the average energy consumption by 76.38\% and achieve an average accuracy of more than 98\% in performing video analytics.

\end{abstract}


\maketitle
\pagestyle{plain}
\section{Introduction}
Video cameras are poised to play a pivotal role in providing advanced analytics in smart cities. Although cameras today are used for surveillance purposes and are required to be in ``always-on" mode, video-based analytics go beyond surveillance and offer rich analytics such as business intelligence, improved operations, environment conservation, and infrastructure management~\cite{ananthanarayanan2017real}. Recent estimates indicate that millions of cameras are deployed in very diverse environments, and operating cameras in always-on mode unnecessarily increase the energy footprint and cost (including electricians to install power supply lines). Moreover, one-fourth of the cameras are standalone battery-powered, where conserving energy is critical~\cite{cameramarket}; the future trend is for standalone cameras to be even more prevalent, given the ease of installation.

An approach to reducing energy footprint is to deploy low-powered camera devices. However, such methods may still have a high energy footprint as processing large videos is computationally expensive. Another effective way to reduce energy consumption is to operate in standby mode in conjunction with using energy-efficient devices. In standby mode, the device suspends all operations and transitions into a low-power mode. This allows the device to consume minimal energy, enough to respond to any wakeup event. At the same time, this also reduces the amount of generated data, reducing the overall computational burden. Many appliances today switch to standby mode when not in use~\cite{wang2020standby}. Studies show that these can lead to significant energy savings over time~\cite{friedli2016energy}. But, such systems require an external signal to switch it to standby mode.  As such, recent efforts have investigated techniques to reduce energy by switching to standby mode based on device usage prediction~\cite{lee2013automatic}.

We note that most cameras for video analytics need not necessarily operate in a continuous-on mode, and thus, there is significant potential in reducing energy use~\cite{friedli2016energy}. 
For example, 
a parking bay's video analysis can determine vacant spots, but such systems need not always be on as long as it provides parking information in a timely manner. If the parking lot is near full, 
a driver may need assistance in locating a spot, as the average driver spends 17 hours per year searching for vacant parking bays~\cite{17hoursparking}. On the other hand,  if the parking lot is near empty, a vacant parking bay's exact location may be irrelevant since it should be easy to find a spot to park.  As such, if we turn off parking video-analytics when parking space is ample, we can tradeoff utility for energy. This is the key idea we study in our work.

Our focus is to develop a reinforcement learning (RL) technique to learn a standby management policy that increases the overall energy savings while retaining the utility of a parking-based video analytics platform. We assume a parking management system that identifies available parking bays using cameras~\cite{valipour2016parking}. Studies have shown that deep learning algorithms can locate occupied and vacant spots with high accuracy~\cite{valipour2016parking,amato2017deep}. However, prior work does not consider the utility-energy tradeoff and assumes these systems to be always operational. Our key insight is that we can relax this notion of utility; that is, the exact location of parking spaces is less important when there is ample parking, and parking analytics can be turned off to save energy. On the other hand, if the parking is full or near full, we expect the analytics to be operational to provide parking occupancy information to users. 

Rule-based standby policy exists that activates the system based on some fixed time. For example, we can program the system to operate only during the day time between 9 am to 5 pm. However, as we show in our analysis, such a policy may not be optimal in saving energy. In contrast, by formulating an RL problem, our proposed approach learns the rules and adapts to the parking patterns to operate the camera. Our approach can learn the parking lot's dynamics by (autonomously) interacting with the environment through reward signals. In doing so, we develop a system that can provide video analytics when needed but learns to conserve energy at other times. Our contributions are as follows:
 
\begin{itemize}
\item \textbf{Parking Data Characterization and Analysis}. We analyze parking patterns on a city-scale through an actual parking dataset from the city of Melbourne, Australia~\cite{parkingdataset}, containing parking occupancy from 76 streets 
collected using ground sensors. 
Our studies show streets have high occupancy during weekdays, weekends, or uniformly distributed, depicting variability.  We also find most streets have high occupancy at midday or in the evening and can be clustered as such. 
This analysis points to the need for an adaptive standby policy that can adjust to the various parking patterns.  

\item \textbf{RL Formulation}. We formulate the parking video analytics problem using RL and propose RL-CamSleep, a deep reinforcement learning approach, that can learn an adaptive standby policy even when parking occupancy information is unavailable. Moreover, our formulation provides control knobs, an input parameter to the reward function, to balance the tradeoff between utility and energy. Thus, it provides the flexibility to choose between utility and energy savings.   
\item \textbf{Design and Implementation}. We design RL-CamSleep's system architecture and implement a simulation parking environment. In particular, we built our environment using the OpenAI Gym framework to simulate parking occupancy for multiple streets in a city.

\item \textbf{Evaluation}. 
We examine the performance of RL-CamSleep on parking data from 10 streets and show that it adapted to different parking situations. We also analyze the adaptability of the learned policy on the unseen 66 streets spread across the city.  Our results show that our approach outperforms other baseline techniques and attains 98.65\% accuracy while achieving a 76.38\% energy savings on the city-scale dataset.  
\end{itemize}

\section{Background}
\label{sec:background}
In this section, we provide background on smart cameras and  
parking guidance systems.

\subsection{Smart Cameras}
Smart cameras play an essential role in any video analytics system. A smart camera typically has remote-control capabilities, is connected to the Internet over a wireless interface, and streams the video data over the IP-network to a centralized (Cloud) system (e.g., Google Nest Cam~\cite{googlenest}).
These types of cameras rely on a two-tier cloud-based architecture, where the video is transmitted over the Internet for analytics processing in the cloud.  The two-tier architecture also enables remote access to the cameras over the Internet~\cite{bovornkeeratiroj2020repel}. This is the basic architecture assumed in our work (see Figure\ref{fig:image_processing}). 

Modern smart cameras are equipped with limited processing capabilities that allow for additional functionality
such as video encoding (e.g., H.264~\cite{wiegand2003overview}). This convenient increase in connectivity and functionality increases the overall energy consumption. To reduce energy consumption, smart cameras have a standby mode, wherein image sensors and other computing hardware is put to sleep but ready to "wake up" when it receives an input signal.  
Since image sensors constitute a large proportion of energy consumption in a video camera, standby mode can save significant energy --- as much as 95\% reduction in energy~\cite{likamwa2013energy}.

\subsection{Parking Systems and Energy-Efficiency}

\begin{figure}[t]
\centering
\includegraphics[width=2.7in]{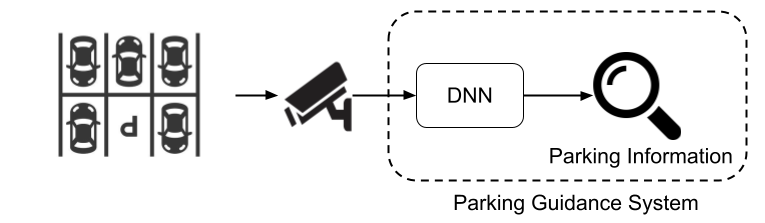}
\caption{Illustrative example of a parking detection analytics platform.}
\label{fig:image_processing}
\end{figure}

Parking Guidance Systems (PGS) that provide drivers with the exact location of vacant spaces can save significant time and costs. Studies show that people often cruise on average 7.8 minutes for a parking spot, which increases traffic congestion in cities~\cite{arnott2006integrated,shoup2011high}. 
Thus, existing PGS mainly adopt either sensor-based or camera-based to identify the exact vacant spot.  In a sensor-based approach, ground sensors are installed in each parking bay, which relay occupancy status in a real-time manner. However, such sensor-based solutions have high installation and maintenance costs, making them cost-ineffective in open areas such as on-street parking areas. As such, camera-based PGS has gained traction,  as they are cheaper to install and also provide high accuracy in identifying the location of vacant spaces. 

With the advancement in deep learning, most camera-based PGS uses deep learning-based techniques to detect parking information. As shown in Figure~\ref{fig:image_processing}, smart cameras relay parking information as a series of individual images or as short video segments (i.e., a batch of consecutive images) to the classification neural network, which then uses inference to extract relevant aspects from the data. We assume a similar setup in our approach. 
However, deep learning-based techniques tend to be computationally intensive and consume much power~\cite{schwartz2019green}. Typically, it requires server-class platforms with GPUs, consuming hundreds of watts. Recent studies have focused on making deep learning networks less power-hungry by designing lightweight architectures~\cite{zhang2018highly}, or specialized hardware accelerators~\cite{park2018energy}. There have also been studies on modifying the inputs (e.g., reducing video frame sampling rate, resizing, resolution) to the neural network to reduce the computing demand~\cite{jiang2018chameleon}. However, since many of these analyses are expected to run continuously, it can result in high costs and energy. 

\section{Problem and Parking Profiles}
\label{sec:data_characteristics}
We begin by describing our problem, and then analyze our city-scale parking dataset to show the spatial and temporal diversity in parking patterns. 

\subsection{Problem}
Our use case consists of a parking video analytics platform that identifies available parking locations for users. This involves a camera transmitting a parking lot's video to a Cloud system, which analyzes the video frames to identify empty parking bays. Our primary goal is to design a policy that activates the camera only when the parking area nears full capacity. 
Fine-grained analytics can provide the exact locations of empty parking bays or the number of spots available when it is at near capacity by continuously monitoring the lot. However, we can tradeoff fine-grained analytics with high overhead for coarse-grained analytics with lower overhead when the parking lot is near empty.  The analytics platform can report to users about parking bays' availability without providing the exact numbers. This can be achieved by temporarily turning off video analytics during periods of high availability (i.e., there are many empty parking spaces). Thus, we seek to create a policy that estimates high and low occupancy periods in parking areas and activates the cameras and video analytics only during high occupancy periods to save energy. 

\begin{figure}[t]
\centering
\includegraphics[width=2.3in]{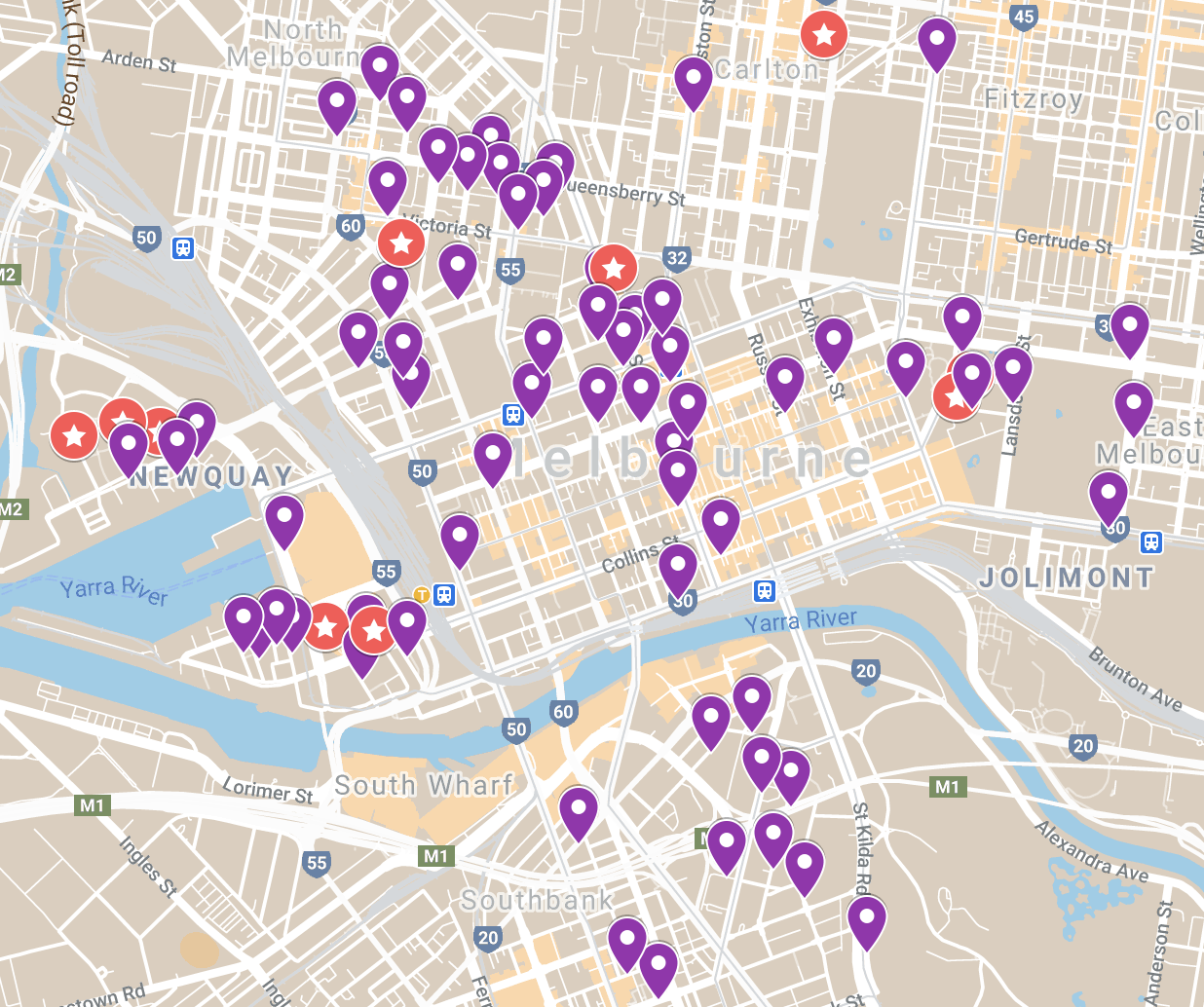}
\caption{Spatial distribution of the streets. The red markers indicate streets in our training dataset, while the purple markers indicate the rest of the city-scale parking dataset.}
\label{fig:street_location}
\end{figure}

\subsection{Dataset and Characterization}
We use the parking dataset from the central business district in Melbourne, Australia~\cite{parkingdataset}, which consists of the arrival and departure of vehicles across 106 streets for 365 days in 2014. 
Since data from other years was not available or complete, we limit our analysis to 2014. However, we believe that our technique and analysis should remain broadly applicable to other periods, given the traffic characteristics shown below.
For our analysis, we consider streets with ten or more parking bays --- a total of 76 out of the 106 streets. 
This is because streets with a lower number of parking bays mostly had high occupancy, with few available parking bays at most times, making it necessary for the camera to be turned on at all times. We discuss this issue in Section~\ref{sec:discussion}. 

We process the data as follows. We first determine the occupancy state (as a percentage of the total) for each street at a one-minute resolution using the arrival and departure information. Next, we divide the occupancy state into three categories (high, medium, and low) based on a given threshold to indicate the parking lot availability level. 
The categories indicate the conditions to activate (or deactivate) the analytics. 
A high occupancy state means there are few parking spots available, and cameras should be turned on to track which parking spots are available. A medium and low occupancy state depicts higher parking spot availability, and thus, cameras could be turned off to save energy and network transmissions. Although, in our analysis, we use 80\% and 60\% thresholds for high and medium, these input parameters can be varied by the user.  Table~\ref{table:data_stat} summarizes the key characteristics of the dataset and the thresholds used. 

\begin{figure*}[t]
        \centering
        \begin{subfigure}[b]{0.19\textwidth}
            \centering
            \includegraphics[width=\textwidth]{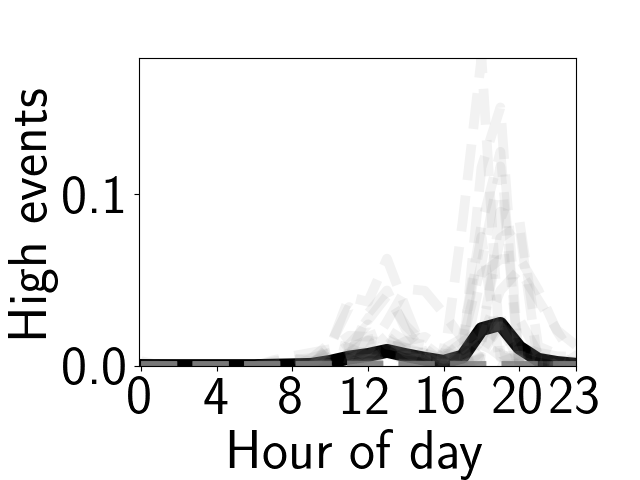}
            \caption[]
            {{\small Bimodal:7pm peak}}    
            \label{fig:bimodal_distr}
        \end{subfigure}
        \begin{subfigure}[b]{0.19\textwidth}  
            \centering 
            \includegraphics[width=\textwidth]{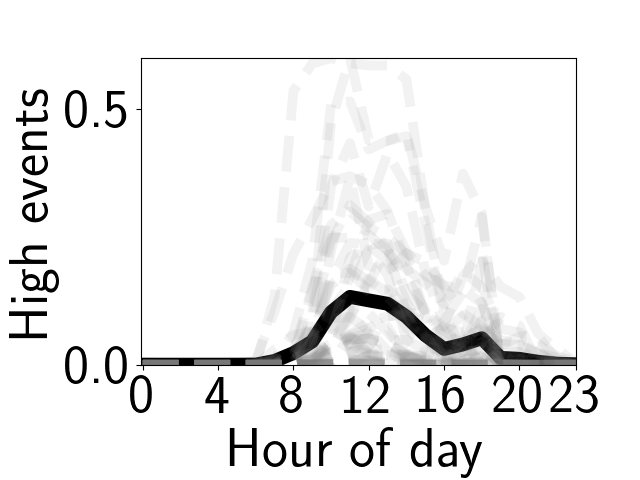}
            \caption[]
            {{\small Bimodal:12pm peak}}    
            \label{fig:unimodal_distr}
        \end{subfigure}
        \begin{subfigure}[b]{0.19\textwidth}   
            \centering 
            \includegraphics[width=\textwidth]{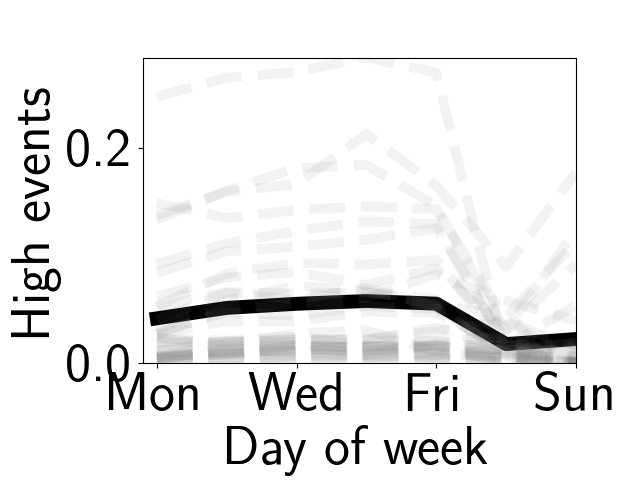}
            \caption[]
            {{\small Weekday profile}}    
            \label{fig:weekday_distr}
        \end{subfigure}
        \begin{subfigure}[b]{0.19\textwidth}   
            \centering 
            \includegraphics[width=\textwidth]{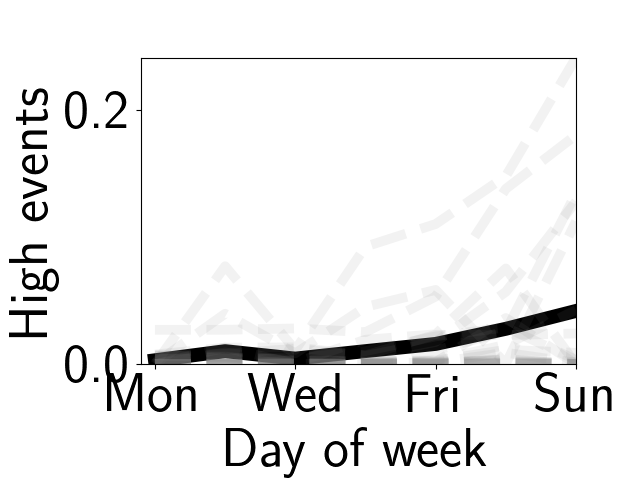}
            \caption[]
            {{\small Weekend profile}}    
            \label{fig:weekend_distr}
        \end{subfigure}
        \begin{subfigure}[b]{0.19\textwidth}   
            \centering 
            \includegraphics[width=\textwidth]{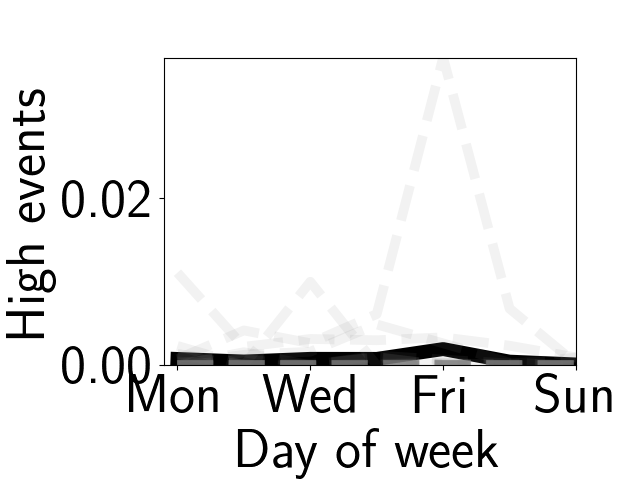}
            \caption[]
            {{\small Uniform profile}}    
            \label{fig:flat_distr}
        \end{subfigure}
        \caption[]
        {Distribution of high occupancy events for different clusters (a-b) the hour of the day (c-e) the day of the week.}
        \label{fig:cluster_distr}
\end{figure*}

\begin{table}[t]
\centering
\footnotesize
\begin{tabular}{|l|l|}
\hline
\textbf{Attribute}                                                                              & \textbf{Value} \\ \hline
\textbf{Number of streets with 10+ parking spots}                                                                      & 76             \\ \hline
\textbf{Total parking bays}                                                                      & 5,070             \\ \hline
\textbf{\begin{tabular}[c]{@{}l@{}}Avg. number of parking bays per street\end{tabular}}   & 66.71 \tablefootnote{Each street may consist of multiple blocks.}         \\ \hline
\textbf{High occupancy threshold}                                        & $\ge$0.8            \\ \hline
\textbf{Medium occupancy threshold}                                       & 0.8$>$ x $\ge$ 0.6            \\ \hline
\textbf{Low occupancy threshold}                                       & $<$ 0.6            \\ \hline
\textbf{Min. \#High occupancy events for any street}                                                       & 0              \\ \hline
\textbf{Max. \#High occupancy events for any street}                                                       & 100,787         \\ \hline
\textbf{\begin{tabular}[c]{@{}l@{}}Avg. \#High occupancy events per street\end{tabular}} & 11,901       \\ \hline
\textbf{\begin{tabular}[c]{@{}l@{}}Total \#events per street\end{tabular}} & 525,600       \\ \hline
\textbf{\begin{tabular}[c]{@{}l@{}}Avg. \% of High occupancy events per street\end{tabular}} & 2.26\%       \\ \hline
\end{tabular}
\caption{Key characteristics of the parking dataset.}
\label{table:data_stat}
\end{table}

\begin{table}[t]
\footnotesize
\centering
\begin{tabular}{|l|c|c|c|c|c|}
\hline
\multirow{2}{*}{\textbf{}}                                  & \multicolumn{2}{c|}{Hourly Distribution}                                                                            & \multicolumn{3}{c|}{Daily Distribution} \\ \cline{2-6} 
                                                            & \begin{tabular}[c]{@{}c@{}}Bimodal\\ (Noon)\end{tabular} & \begin{tabular}[c]{@{}c@{}}Bimodal\\ (7 pm)\end{tabular} & Weekday     & Weekend     & Uniform     \\ \hline
\begin{tabular}[c]{@{}l@{}}Number of\\ Streets\end{tabular} & 44                                                       & 32                                                       & 32          & 21          & 23          \\ \hline
\end{tabular}
\caption{Parking profile clusters  using k-means.}
\label{table:cluster_stat}
\end{table}

Figure~\ref{fig:street_location} shows the location of the streets in our dataset. As shown, the dataset depicts spatial diversity with streets spread across shopping areas, a university, and other points of interest. Next, we analyze the temporal characteristics of high occupancy periods 
to understand when to activate video analytics (i.e., when parking spots are less likely to be available). To do so, we first aggregate the high occupancy periods hourly (and daily), and normalize to range between 0 and 1. Next, we cluster the parking profile using the k-means clustering~\cite{kanungo2002efficient}. 
From our analysis, when clustered on the hourly distribution, two clusters of parking profiles emerged (Figures~\ref{fig:bimodal_distr} and~\ref{fig:unimodal_distr}) and three clusters for the daily distribution.  
Figure~\ref{fig:cluster_distr} shows the hourly and daily high occupancy distribution of all 5 clusters. The black line indicates the average of all the streets, and the grey dashed lines are the individual streets within the cluster. As shown in Figures~\ref{fig:bimodal_distr} and~\ref{fig:unimodal_distr}, the parking spaces are usually ``crowded" during the daytime --- likely due to streets located in the business district. We observe a bimodal distribution with lower parking space availability around 11 am (see Figure~\ref{fig:unimodal_distr}). However, some streets depict a bimodal distribution (Figure~\ref{fig:bimodal_distr}), with a small peak at 1 pm and a large peak at 7 pm --- likely due to proximity to shopping areas and restaurants.  

Figures~\ref{fig:weekday_distr},~\ref{fig:weekend_distr}, and \ref{fig:flat_distr} show the daily distribution of high occupancy periods showing weekday, weekend, and uniform clusters, respectively. The weekday cluster (Figure~\ref{fig:weekday_distr}) is likely from streets located within commercial areas, while weekend clusters (Figure~\ref{fig:weekend_distr}) could be due to restaurants and tourist spots. These 2 clusters account for a majority of the streets --- 42.10\% and 27.63\%, respectively. Finally, the uniform cluster (Figure~\ref{fig:flat_distr}) accounts for 30.26\% of the streets.   

{\bf Key Takeaway}: The parking profiles depict both spatial and temporal diversity. Some streets are busy during the weekdays, others during the weekends. The parking profile also shows hourly variations and indicates that the control algorithm for the cameras must accommodate these variations to achieve both high accuracy and energy savings.

\section{RL-CamSleep Design}
\label{sec:systemdesign}
\begin{figure}[]
\centering
\includegraphics[width=3in]{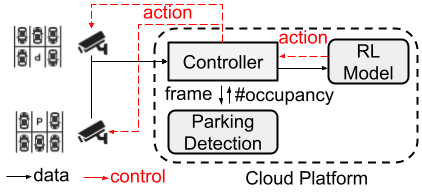}
\caption{Key components in RL-CamSleep system design.}
\label{fig:design}
\end{figure}

The problem of actuating the cameras in an energy-efficient manner can be mapped to making sequential decisions, where the objective is to operate the camera in a manner that achieves high energy savings without sacrificing parking analytics. As we discuss, this problem is well suited for the RL framework used for representing sequential decision-making.  
In this section, we present the key components of RL-CamSleep and present our RL formulation to actuate smart cameras to standby mode in low occupancy periods.   

\subsection{Overview} 
RL-CamSleep approach relies on the observation that the ability to transition between fine-grained and coarse-grained analytics provides the opportunity for gains in energy efficiency. 
Figure~\ref{fig:design} illustrates our approach and highlights how different components work together in decision making. RL-CamSleep has three main components:  \\
{\bf Controller} receives the video frames from cameras and is responsible for actuating cameras. It also coordinates with other components to enable the decision making process.  \\
{\bf Parking detector} detects available parking spots
through object detection algorithms, which can be trained to identify empty parking lots.
Prior work has proposed deep learning based solutions to detect parking bay's occupancy~\cite{amato2017deep}. Our work assumes the availability of parking occupancy information.\\
{\bf RL model} 
receives parking bays information for each street, if available, and decides, for each street, whether to turn on the camera or transition into standby mode.

The controller coordinates with the smart camera and other components as follows. At each time $t$, the controller receives the video frame from cameras that are on, and associates it to the camera identifier. For each received frame, the controller invokes the parking detector to determine the number of parking bays that are occupied in each lot. For cameras that are in standby mode, the controller assumes zero as its occupancy. Next, the RL model is invoked with the parking occupancy information to determine the state of the camera. Finally, on receiving the action from the RL model, the controller actuates the appropriate camera state remotely. We describe each component in detail below.

\subsection{RL Model}
The RL formulation is appropriate because it can learn that a decision at the current time step can have an impact on future results (i.e., if the model decides to put the camera on standby mode, for the next time step it won't have access to occupancy information). Additionally, as described in Section~\ref{sec:data_characteristics}, streets exhibit different parking profiles, and special events (e.g., a parade or a big sale) can cause a decrease in parking availability. RL approaches are adept at making sequential decisions in such uncertain environments~\cite{sutton2018reinforcement}. Moreover, turning on parking video analytics during high occupancy offers itself as a straightforward RL problem, where smart camera actuation represents RL actions and reduced energy footprint and turning on analytics at the right time represents RL rewards.   

In RL, the agent learns to choose an action from its action space that maximizes the sum of discounted future rewards within an environment~\cite{sutton2018reinforcement}. The agent interacts with the environment by selecting an action $a_t = \pi(s_t)$ from a policy $\pi$ and receives a reward $r_t$, where $s_t$ is the state of the environment at time $t$. The goal of the RL is to find a policy that makes the most rewarding decisions. We implement our RL approach as a dueling Double Deep Q-Network (D3QN)~\cite{wang2016dueling, xie2017towards, van2016deep}, a deep learning variant of Q-learning that estimates the action-value function $Q(s, a)$ to learn a policy that maximizes the total reward. The action-value function is also known as Q-value function, where the Q-value represents how useful the action is in gaining some future reward. Below, we define the state space $S$, action space $A$, and reward function $R$ of our RL environment.

\subsubsection{State space} The state $s_t\in S$ characterizes the environment in each time step $t$. The RL model uses the state $s_t$ as input to determine the action $a_t$ for maximizing future rewards. In our work, we characterize the state as follows. 
\\
\noindent
{\bf Observation features ($\mathcal{O}$):} is a vector of current and historical observations of a parking area and is represented as a tuple $(c, o)$, where $c$ represents the state of the camera (i.e., standby or on) and $o$ represents the occupancy percentage of parking bays. The camera state is represented as boolean value $(i.e., c_t \in \{0,1\})$ and the parking occupancy $o_t$ ranges between $[0, 1]$. When the camera is on standby mode, the parking occupancy is set to zero for that time period; when the camera is operating,  $o_t$ is calculated using the parking detector algorithm. The RL environment keeps track of the current and previous $n$ historical observations, which forms a part of the state space $\mathcal{O}_t = \{ (c_{t-n}, o_{t-n}), (c_{t-(n-1)}, o_{t-(n-1)}),...,(c_t, o_t)\}$.  
 
\noindent
{\bf Temporal features ($\tau$):} We also include the time of day, and the day of the week as part of RL-CamSleep's state space, given the insights from Section~\ref{sec:data_characteristics} that show that the streets exhibit temporal characteristics. We encode our temporal values using cyclical representations as we found these techniques to be more robust in our experimentation. We use sine and cosine transformations of the data. 
The final state space $s_t$ is a combination of observations and the temporal features and represented as $s_t = \{\mathcal{O}_t, \tau_t\}.$

\subsubsection{Action space} 
Our action space is simple and consists of two actions: turning the camera on and putting the camera on standby mode (i.e., $a_t = 0$ and $a_t = 1\}$, respectively). The action also affects the video analytics pipeline, as it may reduce the frequency of invoking of the resource-intensive parking detection algorithm. 

\subsubsection{Reward function}
The reward $r_t$ captures the immediate reward an agent receives for an action $a_t$ in state $s_t$. The reward function incorporates two components: the energy consumption and the parking analytics. The energy component captures the energy consumed for operating the camera and executing the parking detection model to determine parking occupancy. Let $e_{1}$ and $e_{2}$ represent the average energy consumed by the camera and the system executing the parking detection model for a duration $d$.  
Then, the penalty for the energy consumption is represented as a combination of both of the components, i.e., $-(e_1 + e_2)d$. Similarly, we also capture the penalty for not operating video analytics during high occupancy periods. Recall that parking video analytics should be operational during high occupancy periods to provide accurate available parking spots. Let $M$ represent the missed instances when the camera was on standby during high occupancy periods. Then, the reward function is.
\begin{equation}
r_t = -[ (\hat e_1 + \hat e_2) \cdot d + \hat w \cdot M] 
\end{equation}
where $\hat e_1, \hat e_2$ represents the normalized penalty for energy consumption, and $\hat w$ is the normalized penalty for missing activating the camera during high occupancy periods. We normalize $e_1, e_2$, and $w$ to transform the components to the same scale. It should be noted that the parameter $\hat w$ controls the tradeoff between energy and utility (i.e., higher the $\hat w$, the less important is energy). We explore the sensitivity of this parameter in our evaluation.   
For simplicity, we do not include communication and other operational costs in our reward function. However, they can be easily incorporated into the reward function as fixed energy costs.

\subsection{Putting it all together}
From our analysis, we find that a key challenge is that high occupancy periods, times when video analytics should be active, are 
rare, at about 2.26\%, which indicates a highly imbalanced dataset.  A vanilla deep Q-network overestimates the action value, leading to a suboptimal policy~\cite{van2016deep}. Thus, RL-CamSleep uses a dueling Double Deep Q-network (D3QN) to prevent overestimation of Q-value and learn the state-value efficiently. 
For more details on the architecture, refer to~\cite{wang2016dueling, van2016deep}.   

The agent estimates the Q-value function by optimizing a loss function iteratively~\cite{mnih2015human}. During this training process, the network is trained by sampling mini-batches of experience $e_t = (s_t, a_t, r_t, s_{t+1})$, which the agent accumulates during the training episodes. This technique is called experience replay and allows the agent to re-use past experiences to train the network. Instead of uniformly sampling the accumulated experience, RL-CamSleep uses a \textit{prioritized experience replay} (PER)~\cite{schaul2015prioritized} that considers more frequently the experiences with a higher difference between the actual and expected reward. 
PER helps the RL model with the unbalanced experiences and results in faster convergence.

\SetKwInput{KwInput}{Input}                
\SetKwInput{KwOutput}{Output}              

\begin{algorithm}[t]
\SetAlgoLined
\KwInput{Parking detector model, $e_1, e_2, w$}
\KwOutput{Actions $a_{c_t}$ for each camera $c$ at step $t$}
  \For{each decision period $t$}{
    \For{each camera $c$}{
        Initialize $o^{c}_{t}$ and $on^{c}_{t}$ with zero;
        \If{$c$ is on}{
            $on^{c}_{t}= $ True\;
			$f^{c}_{t}=$ get\_frame($c$, $t$)\;
			$o^{c}_{t} =$ parking\_detection($f^{c}_{t}$)\; 
        }
        $a^{c}_{t} =$ rl\_model($o^{c}_{t}$, $on^{c}_{t}$, $c$)\;
	    send\_action($a^{c}_{t}$, $c$)\;
    }
  }
 
 \caption{RL-CamSleep controller}
\label{algo:decision_making}
\end{algorithm}

The RL-CamSleep controller uses the RL model and the parking detector for its decision-making process. 
Algorithm~\ref{algo:decision_making} presents the pseudo-code of the controller. The controller uses video frames from cameras that are turned on to determine the occupancy in each parking street. Cameras in standby mode do not transmit video frames, and thus, the occupancy state is unknown. The controller then invokes the RL model along with the occupancy state, if known, to determine the state (on or standby) for each camera. We note that the controller and RL model are not computationally intensive. The RL-CamSleep employs a fully connected neural network with two hidden layers, with 32 and 16 neurons in the first and second layers. One of our strategies for energy saving is to invoke  (computationally intensive) parking detectors only for ON cameras. 

We implemented our RL-CamSleep environment using the OpenAI Gym framework~\cite{openaigym} and the controller using python. The RL environment simulates the energy consumption of the camera and invokes the object detection model. Further, we use the Keras library for D3QN implementation~\cite{chollet2015keras}.  Since year-long (or month-long) parking lot videos were unavailable for training the RL model, we used the numerical parking dataset as a proxy for the parking detector module. Alternatively, we could use prior work on determining parking occupancy~\cite{amato2017deep} as input to our controller. The parking detector module is implemented to replay the parking occupancy for time $t$ and returns the occupancy of the parking lot. Our code and dataset are publicly available at \href{https://github.com/pittcps/rl-parking/}{https://github.com/pittcps/rl-parking/}.

\section{Evaluation Methodology}
\label{sec:evaluation_methodology}
In this section, we briefly describe our training setup, baseline approaches, and metrics for evaluation.

{\bf Training:}
We split our parking dataset with 76 streets into two parts --- 10 streets and 66 streets. 
We use the data from the 10 streets spread across 
the city for training and testing purposes (see Figure~\ref{fig:street_location}).
To do so, we split this dataset into three disjoint datasets as follows: training (50\%), validation (25\%), and testing (25\%). 
We use the rest of the 66 streets for our city-scale validation to analyze the effectiveness of our approach to unseen areas within the city. 

During the training phase, the agent interacts with the environment in episodes, which is the length of the simulation period. We train the model on the data with a one-minute resolution. Further, we train the model for 3200 episodes, where each episode is two weeks long, and the data is randomly selected from a street in the training dataset. 
We use the validation dataset to select a model with the best average reward. We report our result on the test dataset and the rest of the city-scale dataset.

{\bf Baselines:} We compare RL-CamSleep to four methods. We assume all of the following methods send only one frame per second to the analytics service to detect parking spots. 

\begin{itemize}
\item {\bf Optimal:} The optimal method has perfect future knowledge and thus achieves maximum accuracy and energy savings. This cannot be realized in practice and gives an upper bound on the results. 

\item {\bf Naive:} The naive approach simulates a manual configuration approach. This approach assumes the camera is operating for a fixed duration during the day/week and configured accordingly. To calculate the start time and end time for operating the camera, we find the earliest and latest time when the parking lot crossed the high occupancy threshold $\alpha$ in the training dataset for each day and take the minimum and maximum time, respectively. This policy ensures that the camera is always operational during high occupancy periods but switched off during medium and low occupancy. 

\item {\bf Support vector machine (SVM):} The SVM approach simulates a prediction-based approach that uses predicted occupancy to determine actuation. We formulate the problem as a binary classification problem and model the SVM algorithm to predict the camera state using temporal input features (e.g., the time of the day, the day of the week). Each time frame in the dataset is assigned to one of the classes based on whether its occupancy crosses the occupancy threshold  $\alpha$ or not. The goal is to actuate the camera during time frames with higher occupancy than $\alpha$. We do not use parking occupancy as an input feature to SVM because that information is only available when the camera is ON. We used historical known occupancy information but it resulted in poorer performance. 
 
\item {\bf RL-CamSleep-Individual:} This approach simulates a custom RL-CamSleep model for each street. 
In this approach, we train the model report the results from the same street used in training. This baseline allows us to evaluate whether having custom models for each street gives better results than a global model. 
\end{itemize}

{\bf Metrics:}     
We use two metrics to evaluate our approach. 
\begin{itemize}
    \item {\bf Accuracy} captures whether the camera and video analytics are operational during high occupancy periods. We define accuracy as follows: $acc~=~\frac{tot\_high\_{ON}\times100}{tot\_high\_occ}$, where  $tot\_high\_{ON}$ is the number of high occupancy periods that the camera was on during high occupancy periods (in our case, at one-minute interval) and $tot\_high\_occ$ is the number of periods of high occupancy. Thus, an accuracy of 100\% means that the camera was always operational during high occupancy.   
    
    \item {\bf Energy Savings:} is the percentage of the total duration the camera was in standby mode: $sav~=~\frac{tot\_OFF\times100}{tot\_duration}$, where $tot\_OFF$ is the amount of time the analytics was non-operational (camera on standby) and $tot\_duration$ is the total duration of the simulation.
    We note that putting camera analytics on standby reduces both the energy consumed by the camera and the energy consumed by running the parking detection algorithm.
    Moreover, streets may have multiple cameras; thus, we use percentages to compare their performance. 
\end{itemize}

\section{Experimental Results}
\label{sec:experiments}
\begin{table*}[t]
\centering
\footnotesize
\begin{tabular}{|l|c|c|c|c|c|c|c|c|}
\hline
\multicolumn{1}{|c|}{\multirow{2}{*}{\textbf{Model}}} & \multicolumn{4}{c|}{\textbf{Accuracy(\%)}}                           & \multicolumn{4}{c|}{\textbf{Energy Savings(\%)}}                      \\ \cline{2-9} 
\multicolumn{1}{|c|}{}                                & \textbf{Average} & \textbf{Min} & \textbf{Max} & \textbf{Std. Dev.} & \textbf{Average} & \textbf{Min} & \textbf{Max} & \textbf{Std. Dev.}\\ \hline
\textbf{SVM}                                          & 90.66          & 79.12      & 98.71  & 7.6          & 67.59          & 54.83      & 88.48    &9.7 \\ \hline
\textbf{Naive}                                           & 95.74          & 78.64      & 100  &7.6               & 52.91          & 37.49      & 66.66   &8.3   \\ \hline
\textbf{RL-CamSleep-Indi}                         & 98.95          & 98         & 99.85  &0.6           & 76.38          & 62.76      & 90   &8.1      \\ \hline
\textbf{RL-CamSleep}                                         & 99.08          & 97.82      & 99.60    &0.5            & 73.99          & 63.76      & 87.61   & 6.4   \\ \hline
\textbf{Optimal}                                         & 100          & 100      & 100  & 0          & 96.93          & 89.94      & 99.13     &2.9 \\ \hline
\end{tabular}
\caption{Comparison of RL-CamSleep with other baseline methods on test data.}
\label{table:accuracy_energy_Saving_standard}
\end{table*}

\begin{figure}[t]
\centering
\includegraphics[width=3.4in]{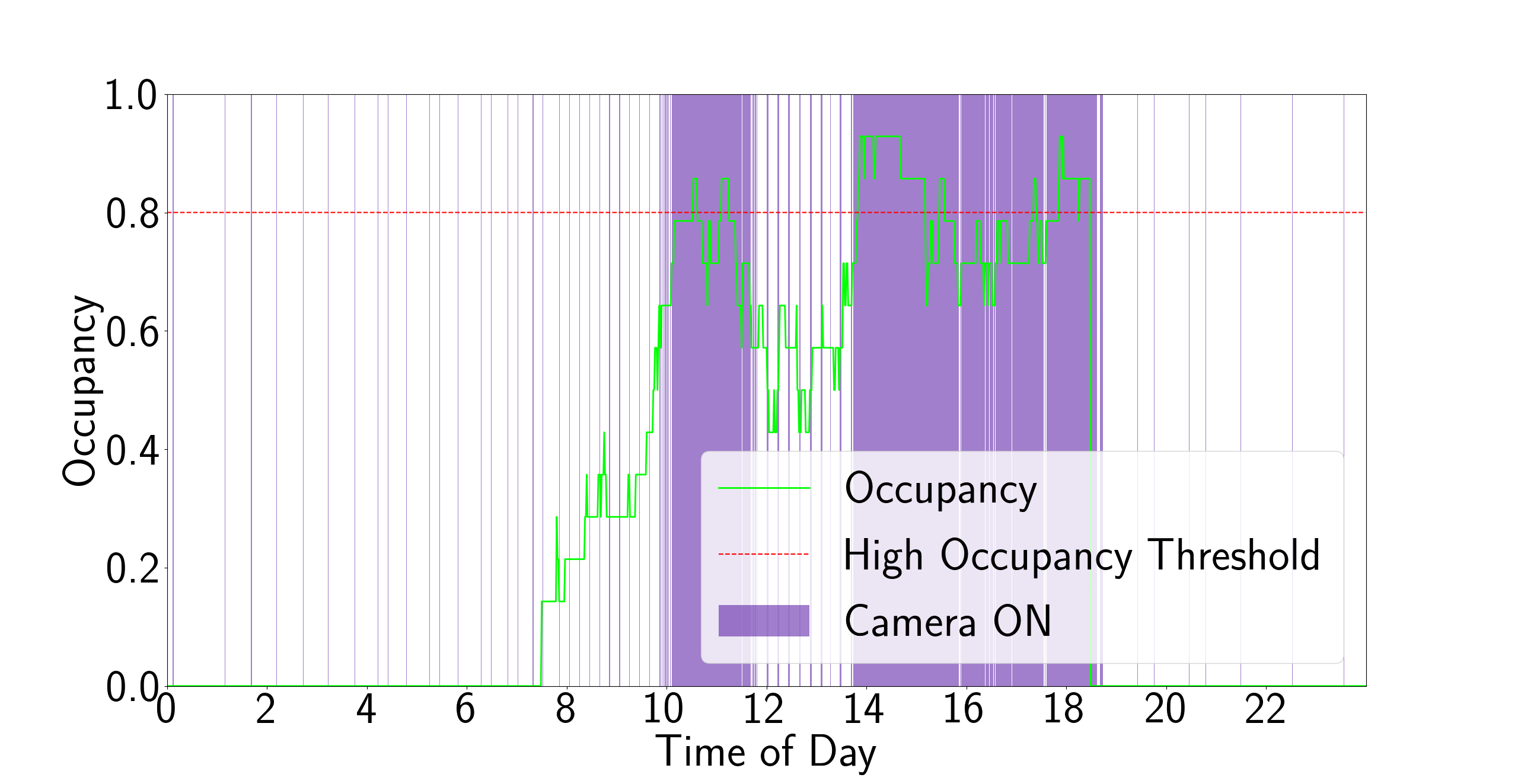}
\caption{RL-CamSleep actions as parking occupancy varies over the day. The figure shows the agent learns an adaptive policy that checks intermittently during low/medium occupancy and activates camera during high occupancy periods.}
\label{fig:one_day_policy}
\end{figure}

In this section, we analyze the performance of RL-CamSleep and compare it to the baseline methods discussed above.

\subsection{Performance comparison}
\label{sub:performance}

We first compare the accuracy of RL-CamSleep with other baseline methods (see Table~\ref{table:accuracy_energy_Saving_standard}). 
Our results show that both Naive and SVM algorithms achieve an average accuracy of 95\% and 90\%, respectively. However, RL-CamSleep outperforms both Naive and SVM methods, achieving an accuracy of 99\%. This shows that the RL-CamSleep successfully learns an adaptive policy to turn on the camera during high occupancy periods. We note that the Naive policy does not perform as well because it is not adaptive. In our evaluation, we fixed the operational time of the camera based on the training dataset. For streets with no change in parking profiles, the Naive method achieves 100\% accuracy. However, our results show that parking profiles change over time, indicating that a fixed policy may not be ideal.  
It is worth noting that the standard deviation of Rl-CamSleep is 15 times lower than the SVM and Naive approaches. This indicates that the RL-CamSleep performance is consistent across the streets. Finally, the RL-CamSleep model achieves similar accuracy to RL-CamSleep-Individual models (trained for each street).

For energy savings, RL-CamSleep performs significantly better than Naive and SVM, achieving 74\% accuracy compared to 53\% and 68\% energy savings, respectively. Furthermore, we observe that both Naive and SVM methods turned on the camera even during low and moderate occupancy periods, resulting in much lower energy savings. This indicates the ability of RL-CamSleep learns a policy that can make decisions to actuate the camera in an energy-efficient manner. 
We observe a marginal increase in energy savings (2\%) when using a custom model for each street (as shown in RL-CamSleep-Indi). However, RL-CamSleep-Indi's standard deviation is higher than RL-CamSeep, indication higher variation in energy savings across different streets.

To better understand the performance of RL-CamSleep, we plot the decisions made by the RL agent for a single street over a day. Figure~\ref{fig:one_day_policy} illustrates the agent's behavior. As shown,
the agent learns to switch on the analytics when the occupancy is above the threshold.  Observe also that the agent learns to actuate the camera for brief periods to get ground truth videos during low occupancy periods. However, this behavior is less frequent at night than during the day resulting in higher energy savings. As we show, this behavior allows the agent to adapt to different scenarios, especially situations where the parking distribution is distinctly different from those in the dataset. 

\textbf{Summary:} \textit{Our results show RL-CamSleep outperforms all baseline models in terms of accuracy and achieves high energy savings. In particular, RL-CamSleep achieves an average accuracy of 99\% and energy savings of 74\%. Our analysis of the agent behavior shows RL-CamSleep learns an adaptive policy that actuates the camera periodically. }

\subsection{Model adaptability}
\label{sub:adaptability}

\begin{figure}[t]
\centering
\includegraphics[width=3.2in]{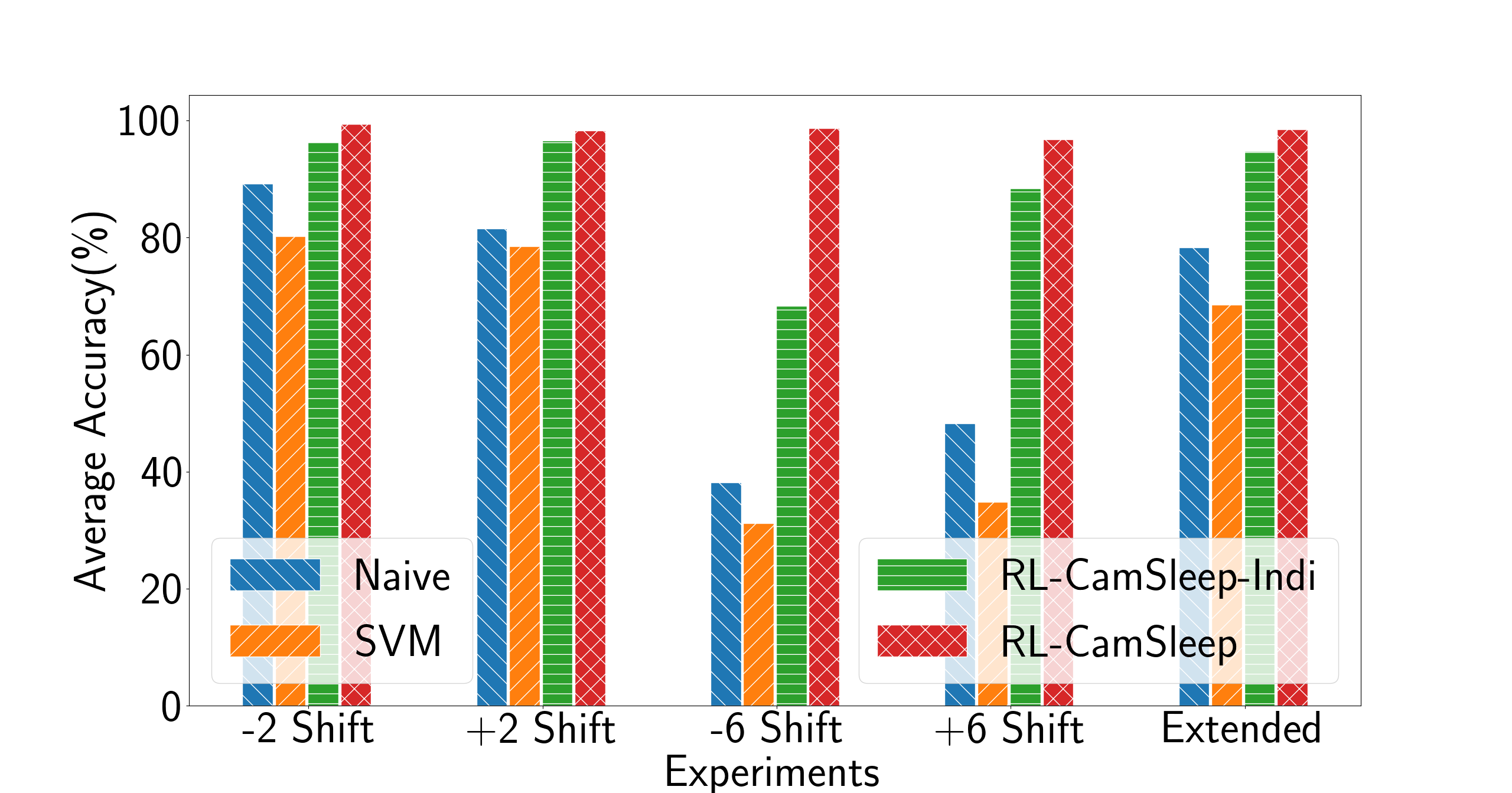}
\caption{Average accuracy on the adaptability dataset.}
\label{fig:adapt_average_accuracy}
\end{figure}

\begin{figure}[t]
\centering
\includegraphics[width=3.2in]{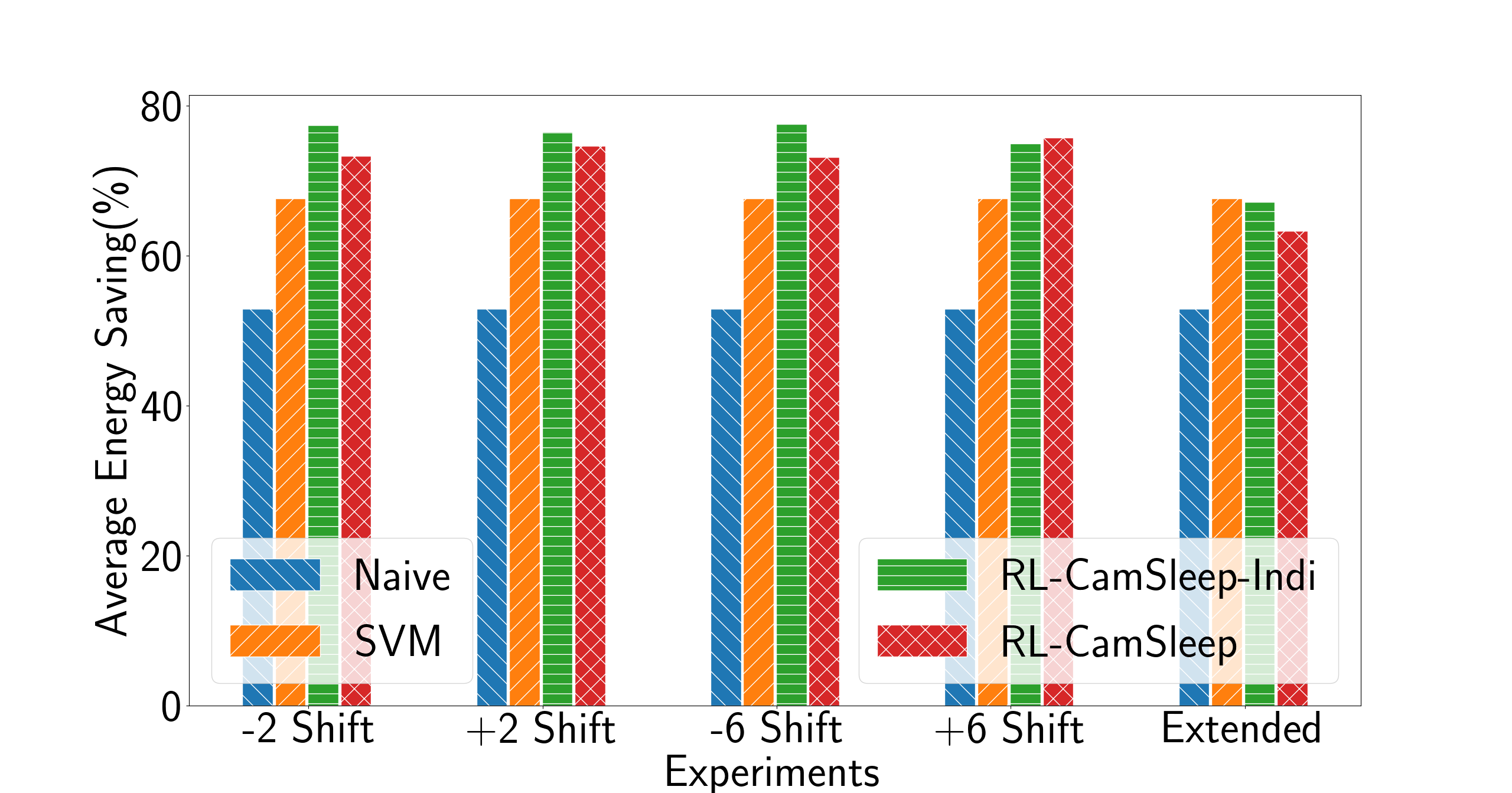}
\caption{Energy savings on the adaptability dataset.}
\label{fig:adapt_average_saving}
\end{figure}

We now evaluate the adaptability of our RL approach to new parking profiles. Such changes in parking profiles may occur due to daylight savings, special events, or change in parking regulations. 
To do so, we modify our testing dataset by shifting the parking distribution by $x$ hours in either direction (or both). For example, a ``-2 hour" (or ``+2 hour") shift indicates that we subtract (or add) two hours from all the points, thereby shifting the occupancy periods. We also extend the distribution in both directions and assume the parking activities start at 5 AM and end at 11 PM.
We achieve this by subtracting 2 hours from all data points before midday and by adding 2 hours to all data points after midday and filling the gap by using the occupancy from the previous time window.   
We report the performance of RL-CamSleep on this synthetically modified dataset below.

Figures~\ref{fig:adapt_average_accuracy} and~\ref{fig:adapt_average_saving} shows the average accuracy and energy savings of different approaches on the modified datasets. We observe that Naive and SVM methods' accuracy drops significantly (by 47\% and 56\%, respectively, for the +6 hours case), indicating that these approaches are unable to adapt to dynamic changes in parking distribution. Interestingly, RL-CamSleep dynamically adapts to different parking distributions as it learns to operate the camera for ground truth observations even during low occupancy periods. However, RL-CampSleep-Individual accuracy drops when there is a significant shift in the distribution (e.g., -6 hours). Since it is trained on one street, it doesn't generalize well, resulting in lower adaptability compared to RL-CamSleep.  RL-CamSleep achieves 96.81\% accuracy --- even when the dataset is modified significantly. Figure~\ref{fig:adapt_average_saving} shows that RL-CamSleep achieves higher energy savings compared to Naive, and SVM methods. 

\textbf{Summary:}  \textit{RL-CamSleep performs better than other baseline methods and can adapt to different situations even when the parking distribution changes. For example, even when the parking distribution is shifted by more than 6 hours, it achieves an accuracy of 97\%. }

\subsection{City-scale analysis}
\label{sub:generalization}

\begin{figure}
        \centering
        \begin{subfigure}[b]{0.23\textwidth}
            \centering
            \includegraphics[width=\textwidth]{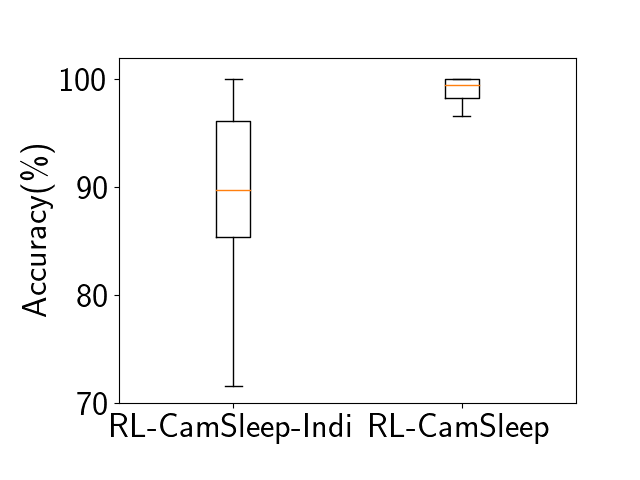}
            \caption[]%
            {{\small Accuracy}}    
            \label{fig:city_accuracy}
        \end{subfigure}
        \begin{subfigure}[b]{0.23\textwidth}  
            \centering 
            \includegraphics[width=\textwidth]{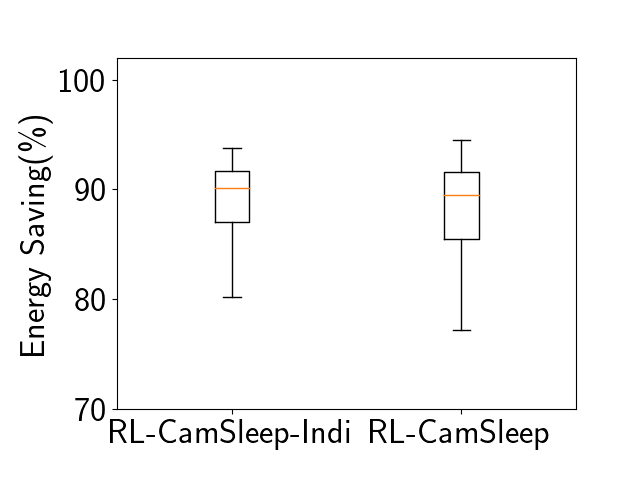}
            \caption[]%
            {{\small Energy Savings}}    
            \label{fig:city_energy}
        \end{subfigure}
        \caption[]
        {Performance on city-scale parking dataset.
        }
        \vspace{-0.1in}
        \label{fig:city_results}
\end{figure}

We now evaluate RL-CamSleep's on the rest of the city-scale parking dataset to show that our approach is scalable to other parts of the city.
The city-scale dataset consists of 66 additional streets, not part of the training or validation dataset. This dataset consists of streets with different high occupancy densities, which means some streets have a high number of high occupancy periods, some streets have a low number and a few streets do not have any high occupancy periods at all. We also compare RL-CamSleep's performance to RL-CamSleep-Individual. To do so, we run each of the ten individual models in RL-CamSleep-Individual on the parking data for the entire year and report the average results.

Figure \ref{fig:city_results} shows the average accuracy and energy savings of RL-CamSleep and RL-CamSleep-Individual. The figure shows that RL-CamSleep performs well even on unseen parking distribution with 98.65\% average accuracy. We also note that RL-CamSleep outperforms RL-CamSleep-Individual by $9.66\%$, indicating that training using data from multiple streets helps in generalization while retaining high accuracy. We also observe that RL-CamSleep has a small standard deviation (2.4\%) compared to RL-CamSleep-Individual, indicating consistent performance. 
RL-CamSleep also achieves an average energy savings of 87.89\% (see Figure~\ref{fig:city_energy}). Although RL-CamSleep-Individual delivers an additional average energy savings of $0.92\%$, this gain comes at the cost of reduced accuracy. The high energy savings from RL-CamSleep on this dataset shows that RL-CamSleep can perform well on different parking distribution and save a substantial amount of energy.

Recall that the city-scale dataset depicts more variation in parking distribution than the streets used in the training dataset. It also includes streets with uniform occupancy distribution and they do not conform to weekday/weekend patterns. We ran RL-CamSleep on streets that belong to the uniform distribution cluster and found that RL-CamSleep performs well even in these unseen streets. In particular, RL-CamSleep shows a 99.79\% average accuracy and 91\% energy savings in this uniform distribution cluster.

\textbf{Summary:}
\textit{Our results indicate that RL-CamSleep achieves an accuracy of 99\% and 88\% in energy savings on city-scale dataset. This shows that RL-CamSleep approach is scalable and can adapt to different occupancy patterns across the city. }

\subsection{Sensitivity analysis}
\label{sec:sens-analysis}

\begin{figure}[t]
\centering
\includegraphics[width=3.2in]{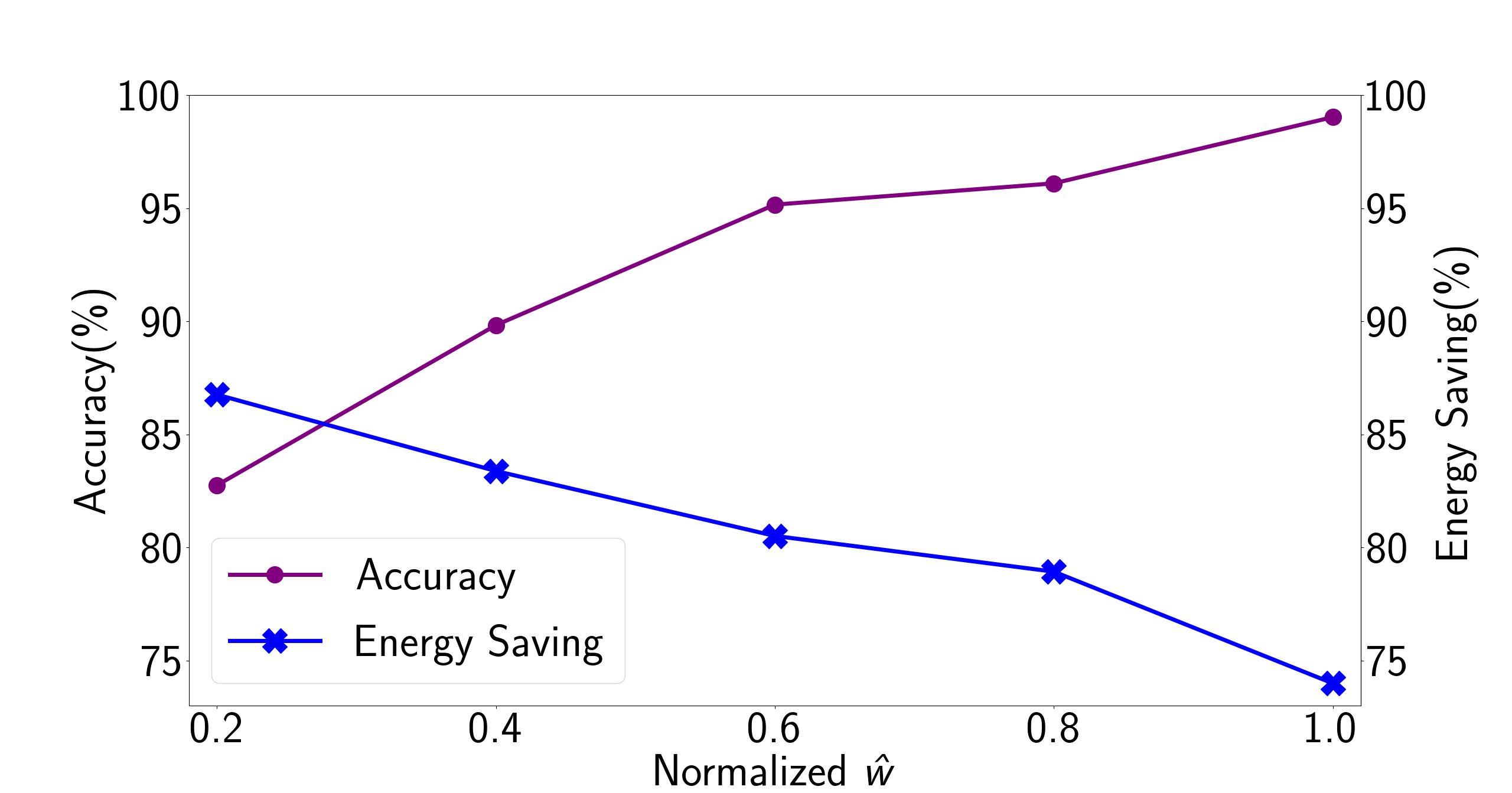}
\caption{Sensitivity analysis for different $\hat w$ values.}
\vspace{-0.1in}
\label{fig:trade_off}
\end{figure}

To understand how we can balance the accuracy and energy tradeoff, we vary the parameter $\hat w$ that controls the preference between energy and accuracy, with a higher value indicating a higher preference for accuracy and vice-versa.  
Figure~\ref{fig:trade_off} depicts the tradeoff between energy and accuracy as we vary the parameter, normalized to the maximum value used in our evaluation. Expectedly, as we increase the value of $\hat w$, the overall accuracy increases and decreases the overall energy savings. This verifies that the agent interacts with the environment using the reward function and learns to balance the two objectives. In particular, we can adjust the $\hat w$ where the results vary between 82.75\% accuracy and 86.76\% energy saving for an energy conservative setting to 99.03\% accuracy and 73.99\% energy saving for a high utility setting.

\textbf{Summary:} 
\textit{ Our results show RL-CamSleep can leverage the utility-energy tradeoff. We can use the control parameter to guide the agent to learn energy conservative or high accuracy policies.}

\subsection{End-to-end performance}

\begin{figure}
        \centering
        \begin{subfigure}[b]{0.23\textwidth}
            \centering
            \includegraphics[width=\textwidth]{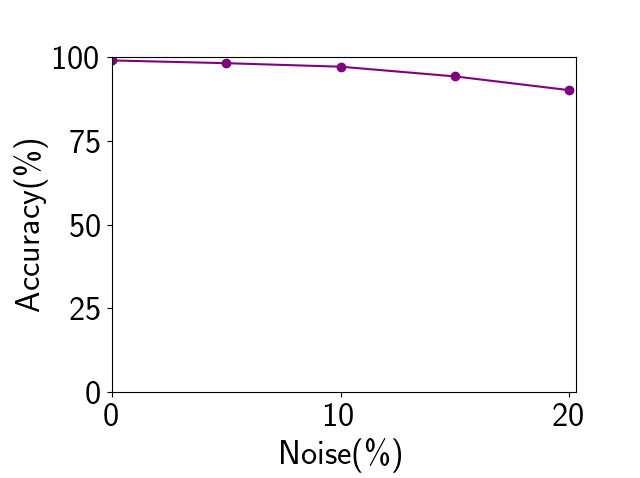}
            \caption[]%
            {{\small Accuracy}}    
            \label{fig:noise_accuracy}
        \end{subfigure}
        \begin{subfigure}[b]{0.23\textwidth}  
            \centering 
            \includegraphics[width=\textwidth]{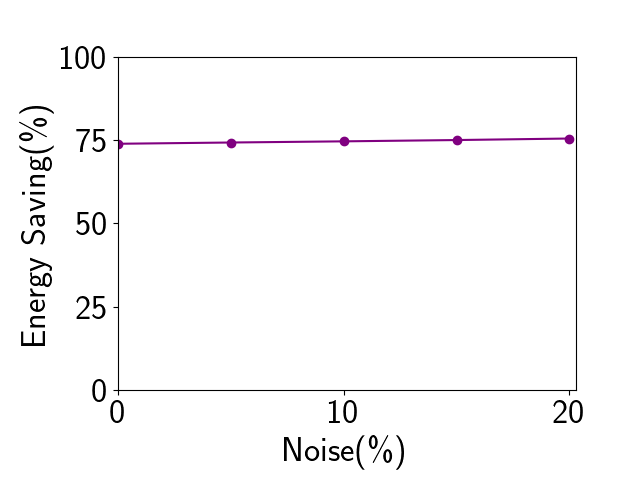}
            \caption[]%
            {{\small Energy Savings}}    
            \label{fig:noise_energy}
        \end{subfigure}
        \caption[]
        {Impact of prediction error on performance.}
        \vspace{-0.1in}
        \label{fig:noise_results}
\end{figure}
So far, we have assumed the parking detector accurately determines the occupancy of a parking area. However, parking detection models may not be accurate, and errors in object detection models (``occupancy sensor'' error) may impact the overall performance. To study how this error propagates, we evaluate the impact of error on the system's overall performance.  To do so, we perturb the original dataset by introducing noise $\delta$ in the parking occupancy value $o$.
We introduce $\delta\%$ noise by randomly selecting a value $x$ from range $[-\delta, +\delta]$ and adding it to the occupancy value i.e., $o_i~=~\max (\min (\frac{(1+x)\cdot o_i}{100} , 1) ,0)$.

Figure \ref{fig:noise_results} shows the impact on accuracy and energy savings as we vary the noise in the testing dataset. The figure reveals that we see a modest decrease in accuracy (of 1.8\%) as we increase the noise from 0\% to 10\%. Prior study has shown that parking detection accuracy range between 96\% and 99.7\%\cite{acharya2018real}. 
This indicates that RL-CamSleep can achieve high accuracy even if the parking detector has a large prediction error. 
We also observe that energy savings stay mostly flat with only a marginal increase of 2.76\% when we increase the noise from 0\% to 30\%.

\textbf{Summary:} \textit{Our end-to-end performance shows RL-CamSleep agent is robust to error in parking detection and can even tolerate a 10\% noise without significant loss in performance.}

\section{Discussion and Future work}
\label{sec:discussion}

The benefit of RL-CamSleep lies in the fact that it can dynamically adapt to various parking scenarios, even when it is different from the distribution seen in the training dataset. As we show, the RL-CamSleep performs well even when some streets have completely different parking patterns. Our approach is adaptable, as reflected in our results of the city-scale simulation. Interestingly, we found that RL-CamSleep learned to check for ground-truth intermittently, even at night, when there is not much activity. This policy makes it more adaptive to dynamic scenarios and we plan to explore the effectiveness of this behavior on other parking datasets. 

A key challenge in applying our technique to real-world use cases is balancing the tradeoff between utility and energy savings. For example, in our analysis, we find that some streets with few available parking spaces always have high occupancy. In such cases,  the cameras tend to be on all the time. This is especially true for streets with low overall parking bays, where the total number of parking is limited (e.g., less than ten). As such, there is little opportunity for energy savings in such streets if utility is prioritized. 

There are also several design considerations and benefits to realizing such energy savings mechanisms in practice. First,  RL-CamSleep can be integrated into parking video analytics and detection systems~\cite{valipour2016parking}, providing both energy savings and cost benefits. Second, although we do not analyze network bandwidth or other Cloud costs, RL-CamSleep significantly reduces the network's traffic and computation (e.g., parking detection) at the cloud server. This is because, in standby mode, the camera stops streaming videos, reducing data and increasing energy savings. We note that this cloud-related cost can easily be incorporated in our analysis if we assume these services are operational only when the camera is operational. Such services can be realized using serverless computing that bill based on a per-request basis. Future work can incorporate such costs as part of the analysis.

\section{Related Work}

The computational cost of executing deep learning models has increased 300,000$\times$ in the past six years~\cite{schwartz2019green}. This has led to a significant increase in energy consumed and will continue and is expected to double every few years~\cite{schwartz2019green}. With future applications relying on such models, it is essential to focus on methods that can reduce their energy footprint. 
Video analytics is one such application that will heavily rely on large neural network models for analysis~\cite{ananthanarayanan2017real}. 
These architectures often become the building blocks for a video analytics application. As such, there have been several studies on video analytics processing to improve the overall energy efficiency of the system~\cite{kang2017noscope, ananthanarayanan2017real}. However, most studies have focused on reducing the latency to enable real-time inference~\cite{kang2017noscope}. Further, since bandwidth is a scarce resource in such live video applications, there have also been efforts to save bandwidth by streaming video at a lower resolution and rate~\cite{ananthanarayanan2017real}. In contrast, our proposed approach is complementary to prior works as we mainly focus on leveraging the tradeoff between utility and energy. Our work showcases how we can save energy, but unlike prior work, we show how RL can be used to learn past events to perform periodic sensing without loss in utility.

Efficient power management schemes can reduce the amount of energy used by using a combination of lower and high power subsystems~\cite{sorber2005turducken}. These systems wake up the more power-hungry resource only when required. 
Prior works have proposed solutions that use dynamic duty cycling based on predicted energy use~\cite{moser2009adaptive}. However, recent work shows that RL can make better energy-efficiency decisions by learning an optimal policy that maximizes the long-term reward~\cite{fraternali2020aces, fraternali2020ember}. In  \cite{lee2013automatic}, authors use device usage and user behavior to manage the standby power consumption automatically. Separately, there have been efforts to use reinforcement learning to improve energy-efficiency~\cite{ding2019octopus, luo2019spoton}. However, prior works do not focus on analyzing energy savings within a parking analytics platform without compromising its utility.  
Moreover, as we show in our evaluation, our work performs better than the prediction-based technique (SVM) in maximizing energy savings and utility. 

RL approaches have also been used for planning purposes and adaptively manage available energy~\cite{xu2020approximate,wei2018reinforcement}. In~\cite{xu2020approximate}, the authors propose a query service for IoT systems that operate within an energy budget. However, the proposed algorithm was evaluated on a 2-week long dataset to provide a counting service. In contrast, our work focus on minimizing the energy consumption while sensing periodically to meet application needs. We evaluated our approach on a year-long city-scale data consisting of 76 streets having over 5000 parking spots. Further, we showed how our approach could be easily transferred to the real world by analyzing the 66 streets not part of the training and testing dataset. To the best of our knowledge, our work is the first to characterize the parking patterns and show potential in energy savings in parking analytics.  We note that there have been studies on parking management systems using neural networks to improve parking spot detection~\cite{valipour2016parking,amato2017deep}. 
Our work is complementary and benefits from advancements in this area.

\section{Conclusion}
Video analytics is gaining traction in many non-surveillance applications. Although prior studies have focused on improving the analytics pipeline, there has been little work on enhancing the platform's overall energy efficiency. In this paper, we considered a parking video analytics platform and proposed RL-CamSleep, a deep reinforcement learning-based technique that can improve the system's overall energy savings while retaining its utility (in the form of accuracy). Our approach is orthogonal to existing work that focuses on improving hardware and software efficiency. We evaluated our approach on a city-scale parking dataset with diverse parking profile patterns. Our city-scale results showed that our RL approach learns a dynamic policy that reduces average energy consumption by 76.38\% and achieves 98\% average accuracy.

\begin{acks}
This research was supported in part by the University of Pittsburgh Center for Research Computing through the resources provided and Google cloud credits for research.
\end{acks}

\balance

\bibliographystyle{ACM-Reference-Format}
\raggedright
\bibliography{paper}

\end{document}